\documentclass[letterpaper, 10 pt, conference]{ieeeconf}  

\IEEEoverridecommandlockouts                              

\usepackage{mathptmx} 
\usepackage{times} 
\usepackage[dvipdfmx]{graphicx}
\usepackage{url}
\usepackage{graphicx}
\usepackage[figurename=Fig.\ ]{caption}
\usepackage{graphics} 
\usepackage{epsfig} 
\usepackage{cite}
\usepackage{color}
\usepackage{booktabs}  %
\usepackage{amsmath}
\usepackage{amssymb}
\usepackage{indentfirst}
\usepackage{algorithm}
\usepackage{algorithmicx}
\usepackage{algpseudocode}
\usepackage{caption}
\usepackage{subcaption}
\newcommand{\minus}{\scalebox{0.75}[1.0]{$-$}}
\newcommand{\figref}[1]{\figurename\ref{fig:#1}}

\title{\LARGE \bf
  Understanding Action Sequences based on Video Captioning \\ for Learning-from-Observation
}

\author{Iori Yanokura$^{1}$, Naoki Wake$^{2}$, Kazuhiro Sasabuchi$^{2}$, Katsushi Ikeuchi$^{2}$, and Masayuki Inaba$^{1}$
\thanks{$^{1}$Iori Yanokura and Masayuki Inaba are with Department of MechanoInformatics,
     University of Tokyo, Tokyo, Japan
     {\tt\small yanokura@jsk.imi.i.u-tokyo.ac.jp; inaba@i.u-tokyo.ac.jp}}%

\thanks{$^{2}$ Naoki Wake, Kazuhiro Sasabuchi and Katsushi Ikeuchi are with Applied Robotics Research, Microsoft, One Microsoft Way,
        Redmond, WA 98052, US
        {\tt\small \{Naoki.Wake,Kazuhiro.Sasabuchi,katsuike\}@microsoft.com
        }}%
}

\begin{document}

\maketitle
\thispagestyle{empty}
\pagestyle{empty}

\begin{abstract}
Learning actions from human demonstration video is promising for intelligent robotic systems.
Extracting the exact section and re-observing the extracted video section in detail is important for imitating complex skills because human motions give valuable hints for robots.
However, the general video understanding methods focus more on the understanding of the full frame,
lacking consideration on extracting accurate sections and aligning them with the human's intent.
We propose a Learning-from-Observation framework that splits and understands a video of a human demonstration with verbal instructions to extract accurate action sequences.
The splitting is done based on local minimum points of the hand velocity, which align human daily-life actions with object-centered face contact transitions required for generating robot motion. Then, we extract a motion description on the split videos using video captioning techniques that are trained from our new daily-life action video dataset. Finally, we match the motion descriptions with the verbal instructions to understand the correct human intent and ignore the unintended actions inside the video.
We evaluate the validity of hand velocity-based video splitting and demonstrate that it is effective.
The experimental results on our new video captioning dataset focusing on daily-life human actions demonstrate the effectiveness of the proposed method.
The source code, trained models, and the dataset will be made available.

\end{abstract}


\section{INTRODUCTION}
Humans can easily understand human actions and imitate them just by watching someone performing them\cite{Kuniyoshi:2003:FromVS}.
It is desired that robots have similar capabilities, which is a challenging task in the field of robotics. 
Considerable research has been done to resolve this 
problem\cite{argall2009survey}\cite{wake2020verbal}\cite{sasabuchi2020task}\cite{nguyen2019v2cnet}\cite{wake2020learning}. 
If robots can understand humans actions, they may acquire new skills without any special programming. 
It is expected, that in the future, in the home environment,
robots will perform daily-life tasks, 
such as serving and cooking, and life support tasks, especially for the elderly.

\begin{figure}[t]
  \centering
  \includegraphics[width=0.48\textwidth]{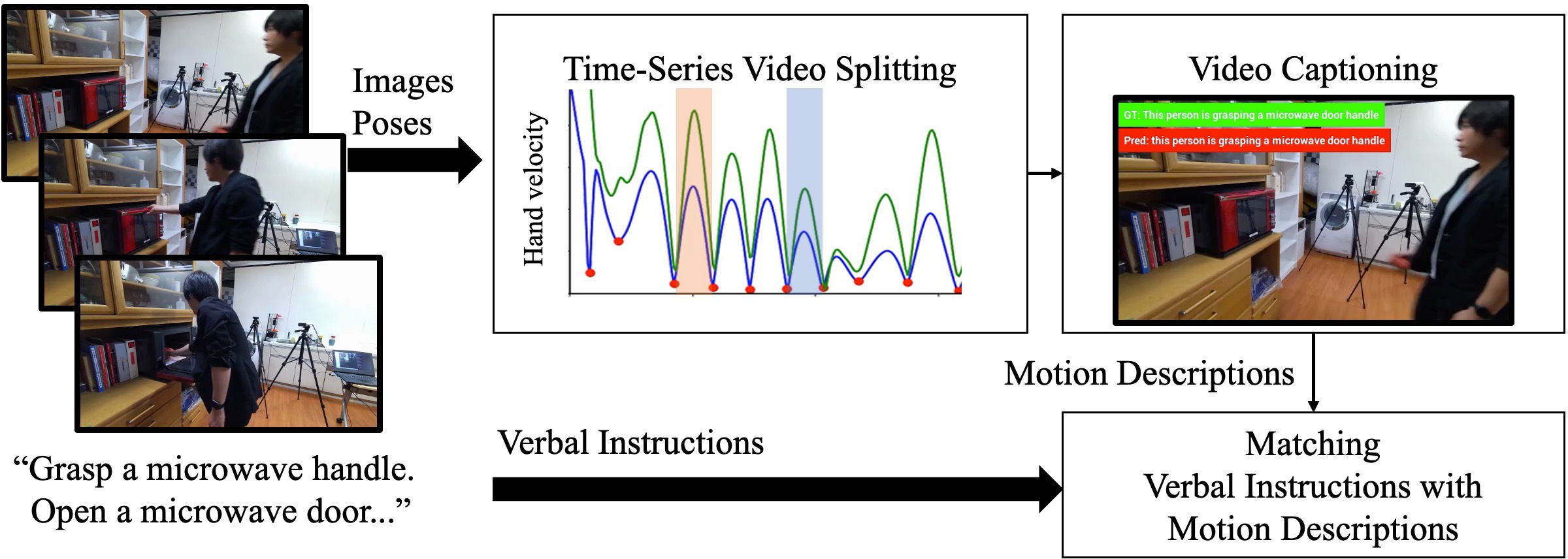}
  \caption{
  An overview of our Learning-from-Observation framework. 
  The framework is composed of three branches: time-series video splitting based on hand velocity, 
  understanding human actions from video captioning, and matching verbal instruction.}
  \label{fig:title}
\end{figure}

In this paper, we argue that three elements are needed to correctly understand humans’ actions: extracting appropriate sections from demonstrations, understanding what a person is doing in each section, and deciding whether a demonstrator intends to perform the action.
The first and second are similar to the end-to-end video captioning problems\cite{densecap}.
With the rise of deep learning, it has become feasible to solve video captioning problems.
Video captioning aims to generate text captions for all events in an untrimmed video.
The task can be decomposed into two problems: detecting event sections and event descriptions.
The former is referred to as temporal action proposal (TAP)\cite{shou2016temporal}\cite{lin2018bsn};
 latter depends on the former.
Although accurate TAP plays an important role in the description of actions from untrimmed videos,
it is difficult to accurately detect proposals.

Recently, research on using video captioning for robots have been conducted\cite{nguyen2018translating}\cite{yang2019learning},
but most works did not focus on detecting accurate sections.
The studies assume that the caption is directly mapped to a robot motion.
However, we argue that the verbal captions are not sufficient to generate a robot motion when several actions are combined,
especially when how the action is performed changes depending on the combination of actions. 
Extracting the exact section and re-observing the extracted video section in detail is important for imitating in such cases.
For example, planning a high-level action, such as manipulating articulated objects with mobile base movement, is difficult without complex modeling of the action sequences,
but the modeling can be relaxed if carefully observed the human action in each extracted section\cite{sasabuchi2020task}.

The third element, deciding whether a demonstrator intends to perform the action, 
is also important for robot teaching because action sequences are complicated and may contain an unintended action.
Such noise can cause robots to misunderstand action sequences.
Recently, there was an attempt to reduce visual noise using verbal instructions in the field of Learning-from-Observation\cite{wake2020verbal}. 
A robot can understand human actions, 
action sequences better if the actions were consistent with how humans verbally describe it.
For example, consider a demonstrator sequentially opening a microwave, placing a cup, 
stopping for a moment to think (noise section), and finally closing the microwave. 
Suppose that a robot understands this as follows: 
the person first opens a microwave, takes a cup, stops for a moment, and finally closes the microwave.
Such misunderstanding is avoided if the demonstrator says, ``open a microwave, place a cup, close the microwave." 
In such a case, the robot can be consistent and understand the action by ignoring the noise section correctly using the verbal instructions.

This paper proposes a Learning-from-Observation framework that splits and understands a video of a human demonstration with verbal instruction.
\figref{title} illustrates the concept of our method.
This paper makes the following contributions:
\begin{itemize}
    \item We discuss the limitation of using video captioning in Learning-from-Observation in terms of accuracy, propose a time-series video splitting method based on hand velocity, and verified the effectiveness of the method.
    \item We present a new video captioning dataset, which focuses on fine-grained daily-life action understanding.
    \item We propose a novel Learning-from-Observation framework, which segments a video and understands demonstrations as a description of the action aligned with the action segments.
    As a result, action sequences understanding can be achieved by matching verbal instructions with the descriptions.
\end{itemize}


\section{RELATED WORKS}
\textbf{Robot Teaching} 
There have been considerable robot teaching studies in the robotics field\cite{argall2009survey}\cite{schaal1999imitation}.
Most of them focused on mimicking the trajectory of a demonstrator's motion\cite{koenemann2014real}\cite{akgun2012trajectories}.
Koenemann et al.\cite{koenemann2014real} presented a system that enables humanoid robots to imitate complex whole-body motions of humans in real time. 
Ijspeert et al.\cite{ijspeert2003learning} proposed an approach for modeling dynamics using dynamic movement primitives by observing human demonstration.
Lee and Ott\cite{lee2011incremental} proposed kinesthetic teaching of motion primitives, encoding trajectories with hidden Markov models. 
However, it is difficult to apply trajectory-based methods to a large number of tasks, such as home environment tasks, which are various and have multiple steps because the training process is designed for a specific task and requires many demonstrations in the real world.

By contrast, there is an approach that maps a human demonstration to a robot via an internal representation, such as with symbolic and geometric constraints, or by understanding with languages of the action.
Arpino et al.\cite{perez2017c} proposed a method that learns multi-step manipulation actions from demonstrations as a sequence of keyframes and a set of geometric constraints. 
Welschehold et al.\cite{welschehold2019combined} proposed action models that enable a robot to consider different likely outcomes of each action and generate feasible trajectories for achieving them. 
Whereas these approaches mainly focus on increasing the feasibility of the action and planning action sequences, 
we focus on understanding action sequences from a human demonstration
video by aligning the video with verbal instructions.

Recently, there has been research on combining human action and natural language.
Plappert et al.\cite{plappert2018learning} proposed a generative model that learns a bidirectional mapping between human whole-body motion and natural language using recurrent neural networks.
Wake et al.\cite{wake2020verbal} encoded human demonstrations into a series of execution units for a robot in the presence of spatio-temporal noise through the combination of verbal instructions.
The former research was aimed at generating motion from language and 
the latter focused on combining verbal instructions with human demonstration, 
we aim to understand human action from a video and extract exact sections for robots.
Moreover, to take advantage of noise reduction with verbal instructions, 
we propose a framework that integrates the whole.

\textbf{Temporal Action Proposal}
TAP is used to predict meaningful proposals from an untrimmed video\cite{shou2016temporal}.
A TAP predicts the start time and end time of an action.
Detecting a meaningful section from a human demonstration video is important because human motions give valuable hints for 
generating a robot motion.
However, the accuracy of detection is still low\cite{lin2018bsn}\cite{lin2019bmn}.
This paper focuses on daily-life actions, such as picking, placing, and manipulating articulated objects. We propose a time-series video splitting method based on hand velocity analysis.
Then, instead of temporal action proposal, 
extracted sections are used as input for video captioning. 

\textbf{Video Captioning}
With the rise of deep learning, 
it is possible to solve the video captioning problem. 
Video captioning has received considerable attention from 
the computer vision and natural language processing communities.
Donahue et.al.\cite{donahue2015long} introduced the first deep learning framework that can interpret an input video as sentences.
Venugopalan et al.\cite{venugopalan2015sequence} proposed an end-to-end sequence-to-sequence model to generate captions for videos from RGB and/or optical flow images.
Zhou et al.\cite{densecap} introduced an end-to-end dense video captioning model to localize (temporal) events from a video and describe each event with natural language sentences. 

Recently, there has also been research using video captioning for robots.
Nguyen et al.\cite{nguyen2018translating}\cite{nguyen2019v2cnet} introduced a method for video-to-command translation for robots.
Yang et al.\cite{yang2019learning} focused on the accuracy of video captioning for robot grasping.
Although the former method produces a multi-step robot command from a human demonstration video, 
it does not detect sections. 
The method does not consider operations with complicated procedures, such as furniture and home appliance operations.
Extracting accurate sections is important for the imitation step of a robot 
because human motions provide valuable hints\cite{sasabuchi2020task}.
The latter approach performs video captioning of what a human grasps in more detail, but it does not consider action sequences.



\section{Problem Formulation}

\subsection{Definition of Actions and Motion Descriptions}
In the Learning-from-Observation field, there are object-centered and body-centered representations for state transition.
The former has been used for manipulation tasks because it does not matter which intermediate poses robots take, 
but rather the face contact transitions the object takes are important\cite{ikeuchi1992towards}\cite{wake2020learning}.
Conversely, the latter has been employed in dance tasks for robots
because how limbs move and how these movements look similar to  human dance are important issues\cite{nakaoka2007learning}\cite{ikeuchi2018describing}.

Because our goal is for robots to generate feasible motions following a human's intent in a home environment, 
we use object-centered representations for describing actions. 
The action is defined as a transition section in which a contact state is switched when a person intentionally tries to operate a target object.

\begin{figure}[htb]
  \centering
  \includegraphics[width=0.4\textwidth]{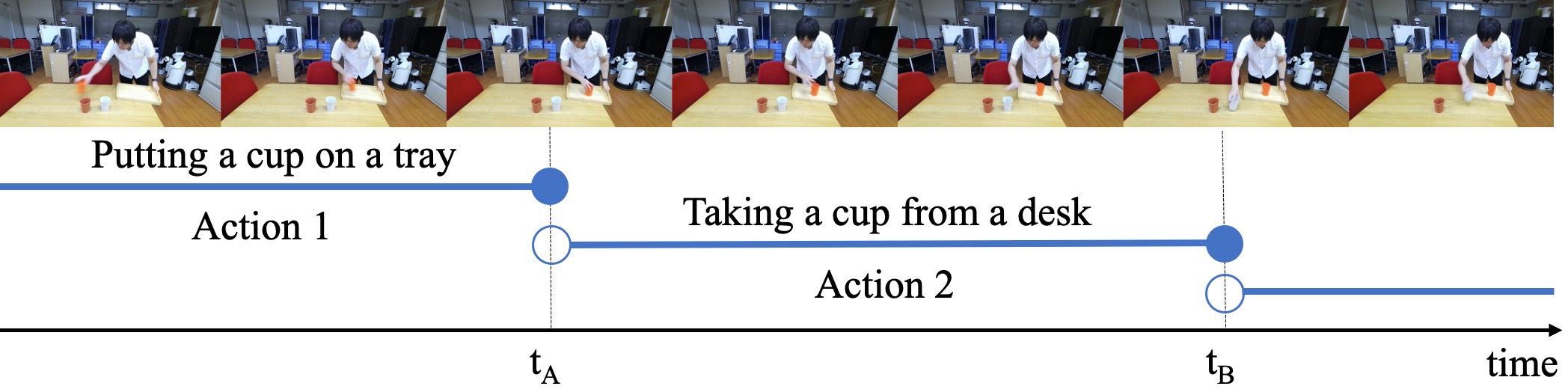}
  \caption{Concept of human actions with motion descriptions.}
  \label{fig:concept-of-tasks}
\end{figure}

In this paper, 
an action is described using natural language, which we call a \textit{motion description}.
\figref{concept-of-tasks} illustrates an example of human actions with motion descriptions. 
Action A is ``putting a cup on a tray'' and Action B is ``taking a cup from a desk''.
At time $t_A$, the demonstrator releases the cup.
This means that the demonstrator's hand and cup are no longer in contact.
After time $t_A$, the demonstrator is going to grasp the cup.
At time $t_B$, the demonstrator grasps the cup.
This is the time at which the hand--cup contact begins.

Grasping and releasing an object causes a state transition between the hand and the object,
and manipulating objects, such as opening a microwave door, causes a state transition between the hand and the object.

\subsection{Definition of the problem for Understanding Action Sequences from video with Verbal Instructions}

Our goal is to extract exact sections for action sequences from video $V = \{v_1, ..., v_L\}$
($L$: the length of frame sequence) aligning with verbal instructions $I = \{i_1, ..., i_N\}$ ($N$: the length of verbal instructions).
Suppose $\pi_\theta$ is a motion description generator from a video (i.e., video captioning).
The $\theta$ is all the learnable parameters for a video captioning model.
The motion description generator aims to generate a sentence (i.e., word sequence) $s = \{y_1, ..., y_T \}$ ($T$: the length of word sequence) to
describe a human action from the video.
\begin{eqnarray}
  \pi_{\theta}(V) = s \nonumber
\end{eqnarray}

Instead of captioning the whole video, we aim to provide captions to sections of the video. Therefore,
we split a video using a video split function as follow:
\begin{eqnarray}
  f_{split}(V) = \{V_1, ..., V_M\} \nonumber
\end{eqnarray}
where $f_{split}$ is a video split function and $M$ is length of the splitted video list. The number of frames of the divided video is 1 or more and $L$ or less.
We then generate motion descriptions for each video section.
\begin{eqnarray}
  S_\theta (f_{split}(V)) & = & \{\pi_\theta(V_1), ..., \pi_\theta(V_M)\} \nonumber \\
    & = & \{s_1, ..., s_M\} \nonumber
\end{eqnarray}
This is the problem of extracting the optimal corresponding sections from motion descriptions $S_\theta$ and verbal instructions $I$.
The overall objective for the problem is formulated as 
\begin{eqnarray}
  L(V, I) & = & \min_{\theta} f_{match} (I, S_\theta (f_{split}(V))) \nonumber 
\end{eqnarray}
where $f_{match}$ means matching function $I$ with $S_\theta (f_{split}(V))$. 
In addition to one-to-one matching, there are cases where multiple elements in $S_\theta$'s are assigned to $i_i (1 <= i <= N)$ in $I$. In this paper, we propose a matching method using dynamic programming for $f_{match}$.


\begin{figure}[htb]
  \centering
  \includegraphics[width=0.4\textwidth]{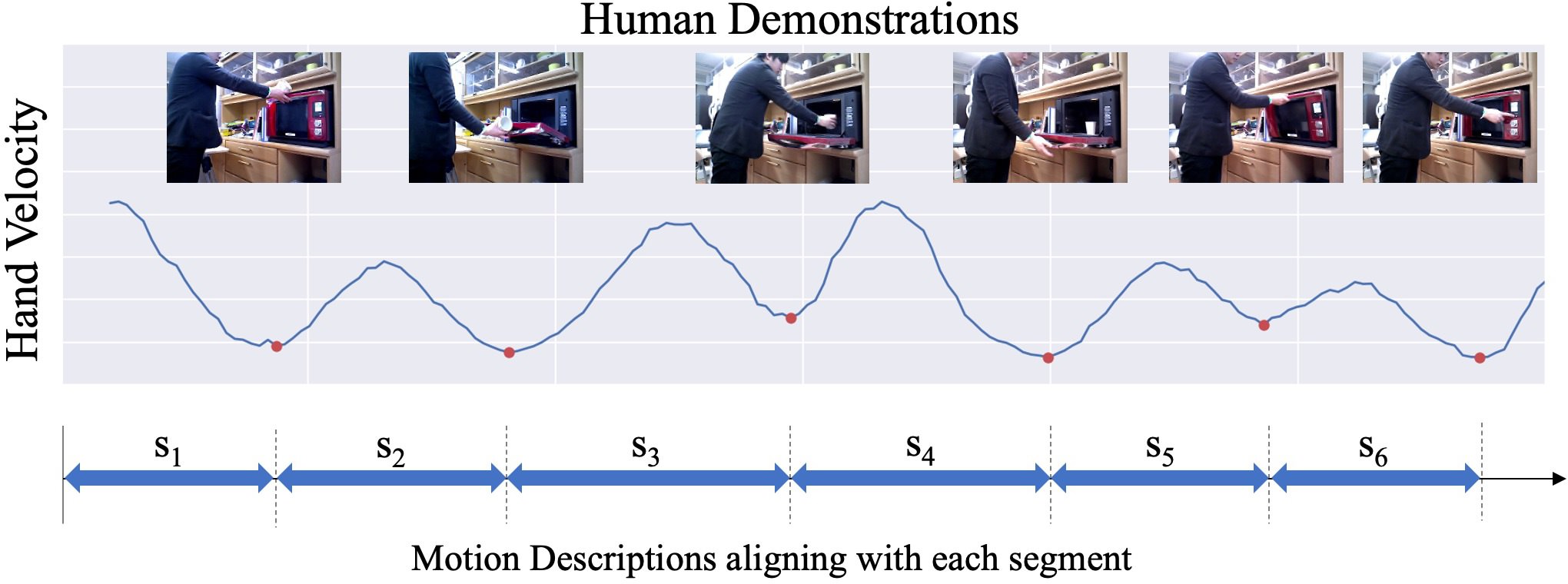}
  \caption{The segmented sequences with motion descriptions. An input video divided by video splitting function $f_{split}$. The motion descriptions are generated from $\pi_{\theta}$. 
  In this paper, velocity-based video splitting and video captioning play a role of $f_{split}$ and $\pi_{\theta}$.}
  \label{fig:problem-formulation}
\end{figure}

\figref{problem-formulation} shows the concept of this problem.
We explain our design of $f_{split}$ and $\pi_\theta$ in the next section.


\section{Implementation}
\subsection{Time-Series Video Splitting based on Velocity}
\begin{figure}[htb]
  \centering
  \includegraphics[width=0.4\textwidth]{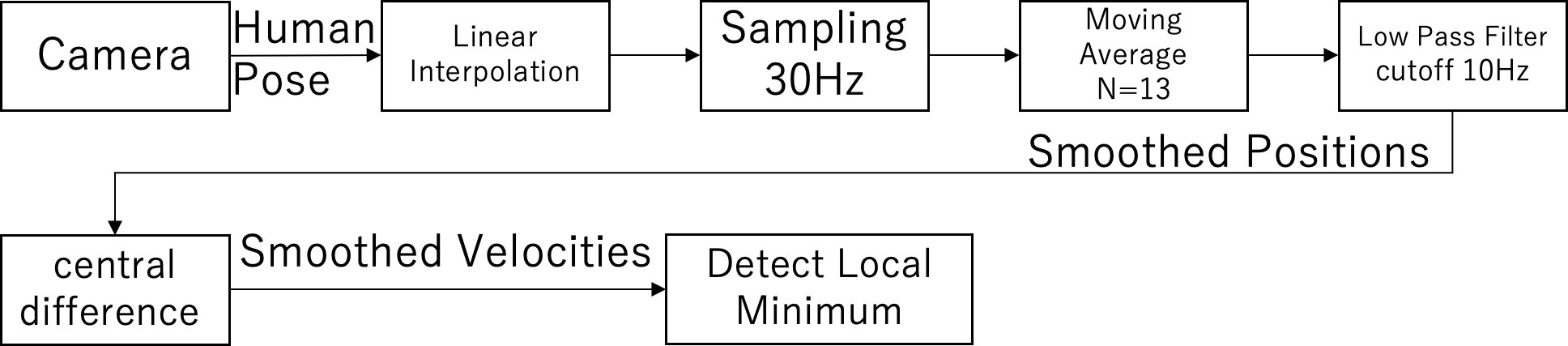}
  \caption{Flow for human motion splitting. Human poses are input from a camera, and the hand velocities' local minimum points are output.}
  \label{fig:flow-of-hand-velocity}
\end{figure}

To realize highly accurate action section extraction, we used a detection method based on human motion characteristics instead of conventional methods, such as TAP.
According to Flash and Hogan\cite{flash1985coordination}, there are two kinds of human limb motions: 
ballistic motion and mass spring motion. The research indicates the end effectors' velocity gives natural indications of the motion segments. 
We design a $f_{split}$ function based on hand velocity from this fact.
Shiratori et al.\cite{shiratori2004detecting} proposed a velocity-based video splitting for dance motions.
Using this knowledge, \figref{flow-of-hand-velocity} indicates the flow for human motion splitting.
RGB-D images and human poses are obtained by a vision sensor\cite{k4a}. 
The human poses are captured in a camera coordinate system.
\figref{flow-of-hand-velocity} illustrates the flow of the proposed video splitting method based on hand velocity.
This flow extracts the local minimum points (times) of the hand velocity from human poses.
To reduce the effect of noise, we use moving average smoothing.

\subsection{Daily-life Motion Description Dataset}
A number of video description datasets have been released\cite{msrvtt}\cite{lea2016learning}\cite{nguyen2018translating}.
However, most of them only provide general descriptions of videos; they do not explain daily-life actions, such as picking, placing, and manipulating articulated objects.
Therefore, we created a new video captioning dataset that focuses on fine-grained daily-life human action understanding.

\textbf{Dataset statistics}
We collected 512 videos of a home environment.
The dataset contains videos of actions such as picking objects, placing objects, arranging objects, and manipulating articulated objects.
We segmented each video and annotated it with captions, resulting in 2985 sections.
Each section has a sentence that describes the human actions.
We used 80\% of the dataset for training and the remaining 20\% for testing.

\subsection{Motion Description using Video Captioning}
As mentioned in section 1, we used video captioning technology as the motion description generator $\pi_\theta$ because the problem's essence is similar to it.
An improvement of video captioning is out of scope in this research.
We used the existing state of the art method\cite{lei2020mart} and 
took the following schemes.

\textbf{Data Preprocessing} 
We added a ``nothing" annotation label for sections representing noise (i.e., non-intended human actions during a demonstration).

For each video, we down-sampled each 0.16 s (6 fps) and extracted the 1-D appearance and optical flow features per frame.
This is higher than the downsampling in existing methods\cite{xiong2016cuhk}\cite{lei2020mart} which use 0.5s (2 fps). We use a much higher downsampling since the actions in home environments are fast.
The video captioning method uses two-stream features: RGB features and optical flow features.
Using these features is relatively common in the fields of action recognition and video captioning\cite{xu2020g}\cite{densecap}\cite{lei2020mart}. 

For the two-stream features, the RGB features were extracted at the Flatten 673 layer of ResNet-200\cite{he2016deep} from the first frame for every 6 consecutive frames,
and the optical flow features were extracted at the global pool layer of BN-Inception\cite{BNInception} from the calculated optical flow between 6 consecutive frames.
We computed the optical flow for our dataset using FlowNet 2.0\cite{flownet2}.

We truncated sequences longer than 80 for videos and 22 for text, and set the maximum number of video segments to 16.
Finally, we built vocabularies based on words that occur at least five times. The resulting vocabulary contains 108 words.

\textbf{Data Augmentation}
To make the model more robust to various variations in the input video, we applied the following data augmentation to our dataset: 

\begin{itemize}
    \item Randomly flip images.
    \item Randomly rotate images.
    \item Add salt and pepper noise to each image.
    \item Randomly change brightness, contrast, saturation, and hue of images.
\end{itemize}

\subsection{Verbal Instructions Matching with Motion Descriptions}
\begin{figure}[htb]
  \centering
  \includegraphics[width=0.2\textwidth]{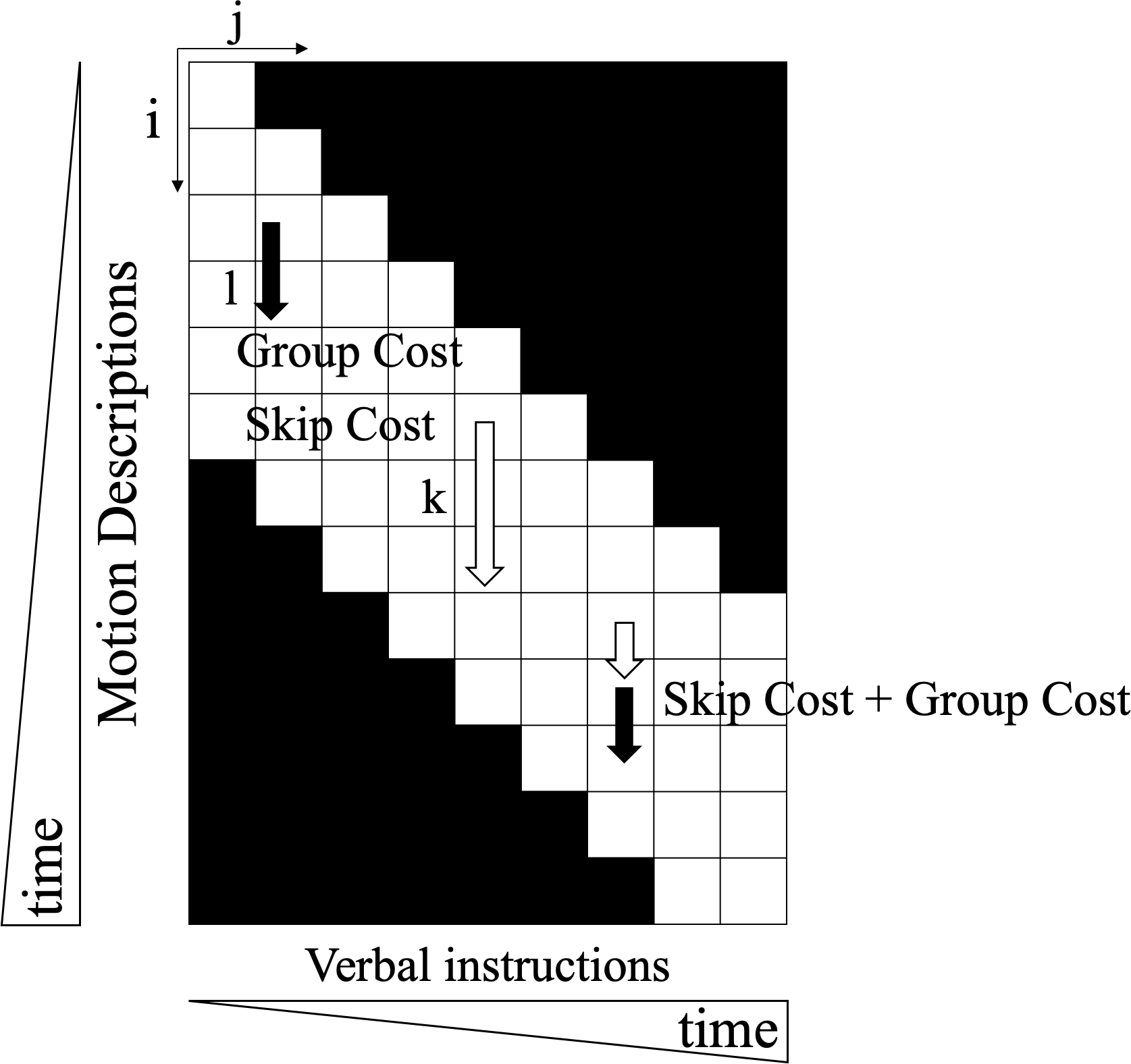}
  \caption{Proposed matching concept. Cells indicate a dynamic programming table for matching verbal instructions with motion descriptions. The blacked-out cells indicate cells that cannot calculate the score.}
  \label{fig:matching}
\end{figure}

In this section, we explain the matching of verbal instructions with the motion descriptions (i.e., video captioning results).
We cast the matching problem as a dynamic programming (DP) problem.
\figref{matching} illustrates the concept of the proposed DP matching.
Each cell indicates a DP score.
The vertical direction indicates motion descriptions $S_\theta = \{s_1, s_2, ..., s_M\}$ ($M$: the length of motion descriptions) and 
horizontal direction indicates verbal instructions $I = \{i_1, i_2, ..., i_N\}$ ($N$: the length of verbal instructions).
The blacked-out cells indicate cells that cannot calculate the score because motion descriptions in those cells cannot be assigned with each verbal instructions.
The matching can be obtained by computing the DP table according to the following formula

\begin{eqnarray}
\lefteqn{ DP[i, j] } \nonumber \\
 & = & min\{DP[i + k + l, j + 1], \nonumber \\
 & & + C_{dist} D(i + k, j) \nonumber \\
 & & + C_{group} GD(i, i + l) \nonumber \\
 & & + C_{skip} k\ \nonumber \\
 | & 0 & <= k < m - i, \nonumber \\
   & 0 & <= l < m - i - k \} \nonumber
\end{eqnarray}

where $D(i, j)$ indicates the distance function between $s_i$ and $i_j$.
$GD(i, i + l)$ represents the accumulated distance of video captioning results from $s_i$ to $s_{i + l}$. 
This has the role of grouping multiple sections together.
Some motion descriptions may not be assigned to verbal instructions, 
which means that $k$ skipped assigning.
$C_{dist}$, $C_{group}$, and $C_{skip}$ are coefficient values.

To get meaning of words, we used pre-trained models trained on Google News\cite{mikolov2013distributed}\cite{blanco2015fast} for Word2Vec.
The distance between $s_i$ and $i_j$ was calculated by word mover's distance (WMD)\cite{kusner2015word} using word embeddings by the Word2Vec.
To make it difficult to match a section captioned ``nothing", 
the distance between $i_j (1 <= j <= N)$ and ``nothing" was set to a constant value $C_{nothing}$.

\section{Evaluation of The Proposed Method}


\subsection{Velocity-based Video Splitting Evaluation}

We evaluate the performance of the velocity-based time-series video splitting. 
Our goal is to show that the hand velocity time-series splitting is a good indicator for dividing daily-life action sequences. To evaluate it, we consider the human action detection as a change point detection problem\cite{kawahara2012sequential}\cite{liu2013change}.

\begin{figure}[htb]
  \centering
  \includegraphics[width=0.3\textwidth]{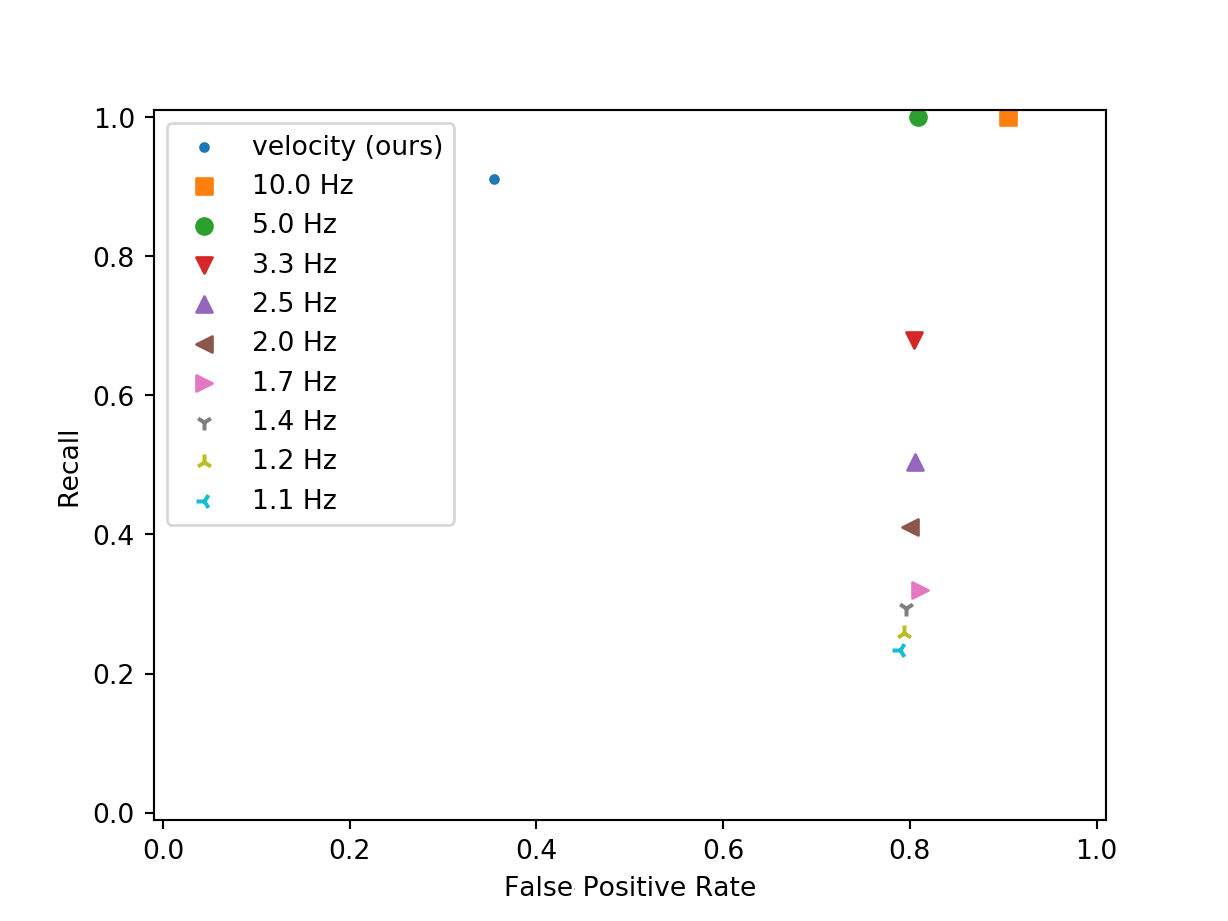}
  \caption{Recall vs false positive rate for action change point detection.}
  \label{fig:recall-vs-fpr}
\end{figure}

We define the true positive rate (recall) and false positive rate as follows: 
\begin{itemize}
    \item Recall: $\frac{n_{cr}}{n_{cp}}$
    \item False positive rate: $\frac{n_{al} - n_{cr}}{n_{al}}$
\end{itemize}
where $n_{cr}$ denotes the number of change points correctly detected, $n_{cp}$ denotes the number of all change points, and $n_{al}$ is the number of all detection alarms.
A detection alarm at time $t$ is regarded as correct if there exists a true alarm at time $t^*$ such that $t \in [t^\ast \minus margin, t^\ast + margin]$.
For comparison, we prepared detection alarms sampled at a certain frequency from the time-series.
We confirm that in most cases, action change points can be tolerated within $\pm0.1$ s.
To evaluate it, we annotated 173 videos, resulting in 1191 change points.
\figref{recall-vs-fpr} illustrates recall vs false positive rate.
The results show that the velocity-based video splitting performs better than the uniform sampling of points. The false positive rate is slightly high, which indicates that over-dividing occurred.
This is not a problem because it can be solved by the matching sections.

\begin{figure}[htb]
  \centering
  \includegraphics[width=0.3\textwidth]{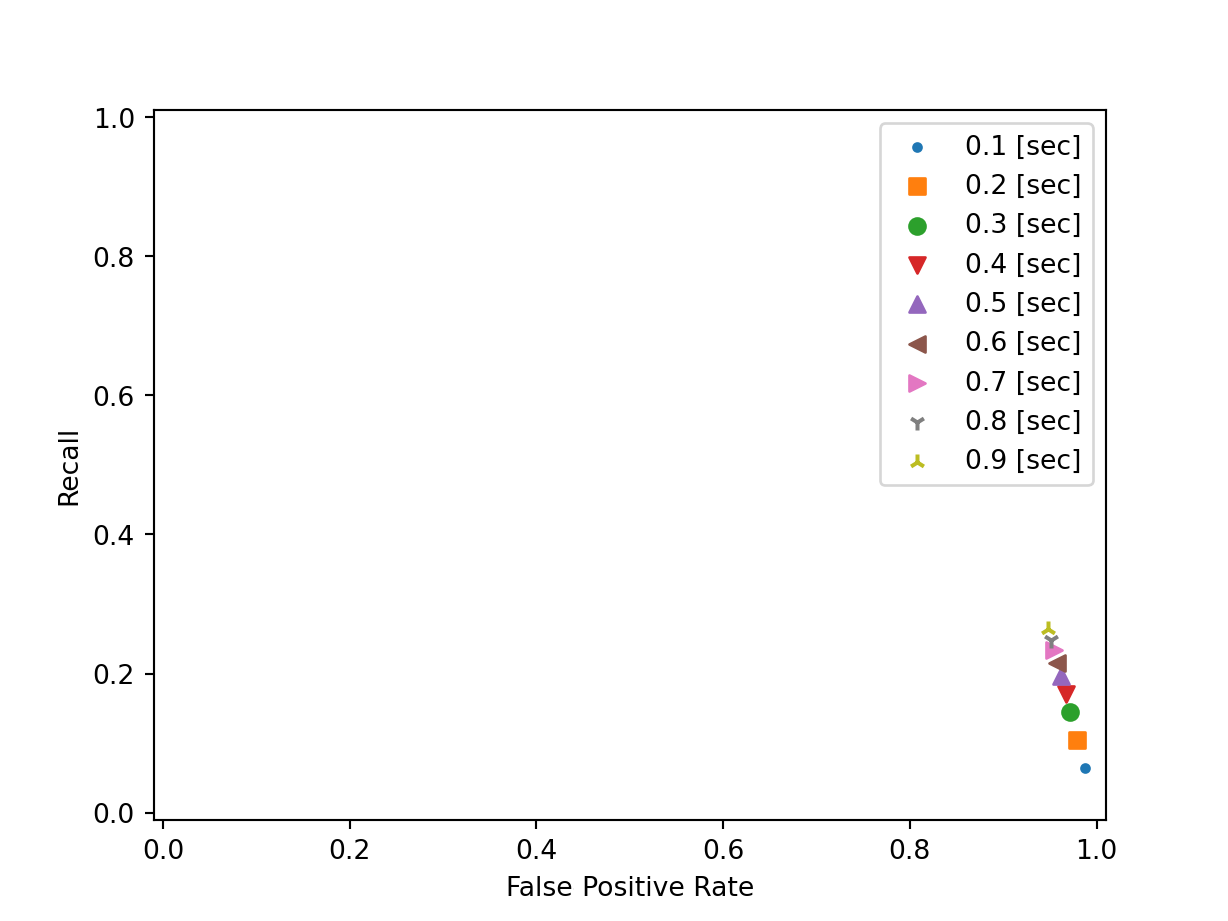}
  \caption{Recall vs false positive rate for end-to-end video captioning.}
  \label{fig:recall-vs-fpr-deep-learning}
\end{figure}

We also evaluated the end-to-end video captioning method\cite{densecap}, whose results are shown in \figref{recall-vs-fpr-deep-learning} 
These results indicate that it is relatively difficult for end-to-end video captioning to extract accurate sections.
Therefore, it is better to conduct hand velocity-based video splitting to divide human demonstrations. 


\subsection{Human Action Description Dataset Evaluation}

We employed K-fold cross-validation\cite{kohavi1995study} to evaluate our dataset.
TABLE 1 shows the results of K-fold cross-validation using video captioning\cite{lei2020mart}.
It can be seen that the variances are small for the language metrics (Blue\_4, METEOR, ROUGE\_L, CIDERr).
These results demonstrate that the proposed dataset is unbiased to the test data.

\begin{table}[htb]
	\centering
	\caption{K-fold cross-validation result of video captioning\cite{lei2020mart} on our dataset.}
	\renewcommand\tabcolsep{4pt}
	\begin{tabular}{rcccc}
		\toprule  %
		&Bleu\_4&METEOR&ROUGE\_L& CIDEr\\
		\midrule
&22.2 & 31.0 & 41.3 & 26.3 \\
&21.1 & 30.3 & 39.2 & 25.4 \\
&23.1 & 30.0 & 38.5 & 20.6 \\
&21.8 & 29.6 & 39.1 & 21.7 \\
&22.6 & 30.4 & 42.1 & 23.4 \\
		\bottomrule  %
Mean&22.2 & 30.2 & 40.0 & 23.5 \\
Variance&0.46 & 0.20 & 1.96 & 4.60 \\
			\bottomrule  %
	
	\end{tabular}
	\label{table:k-fold}
\end{table}

\begin{table}[htb]
	\centering
	\caption{Matching results on testing set of our dataset.}
	\renewcommand\tabcolsep{4pt}
	\begin{tabular}{rcccccc}
		\toprule  %
		Method & 0.5 & 0.75 & 0.95 \\
		\midrule
    Uniform sampling 0.5 & 0.15 & 0.01 & 0.00 \\
    Velocity-based splitting (ours) & 0.59 & 0.57 & 0.56 \\
		\bottomrule  %
	\end{tabular}
	\label{table:matching}
\end{table}


\begin{figure*}
  \centering
  \includegraphics[width=0.9\textwidth]{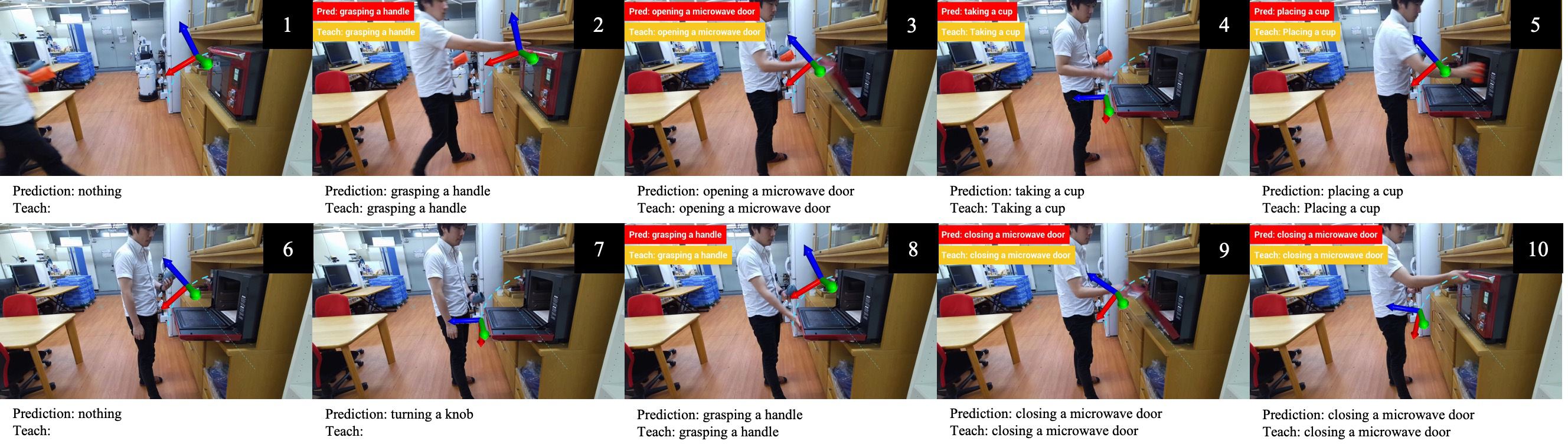}
  \caption{
    The person opens a door, takes a cup, puts the cup into a microwave, and closes the door.
  }
  \label{fig:microwave-operation}
\end{figure*}


\begin{figure}[htb]
  \centering
  \includegraphics[width=0.3\textwidth]{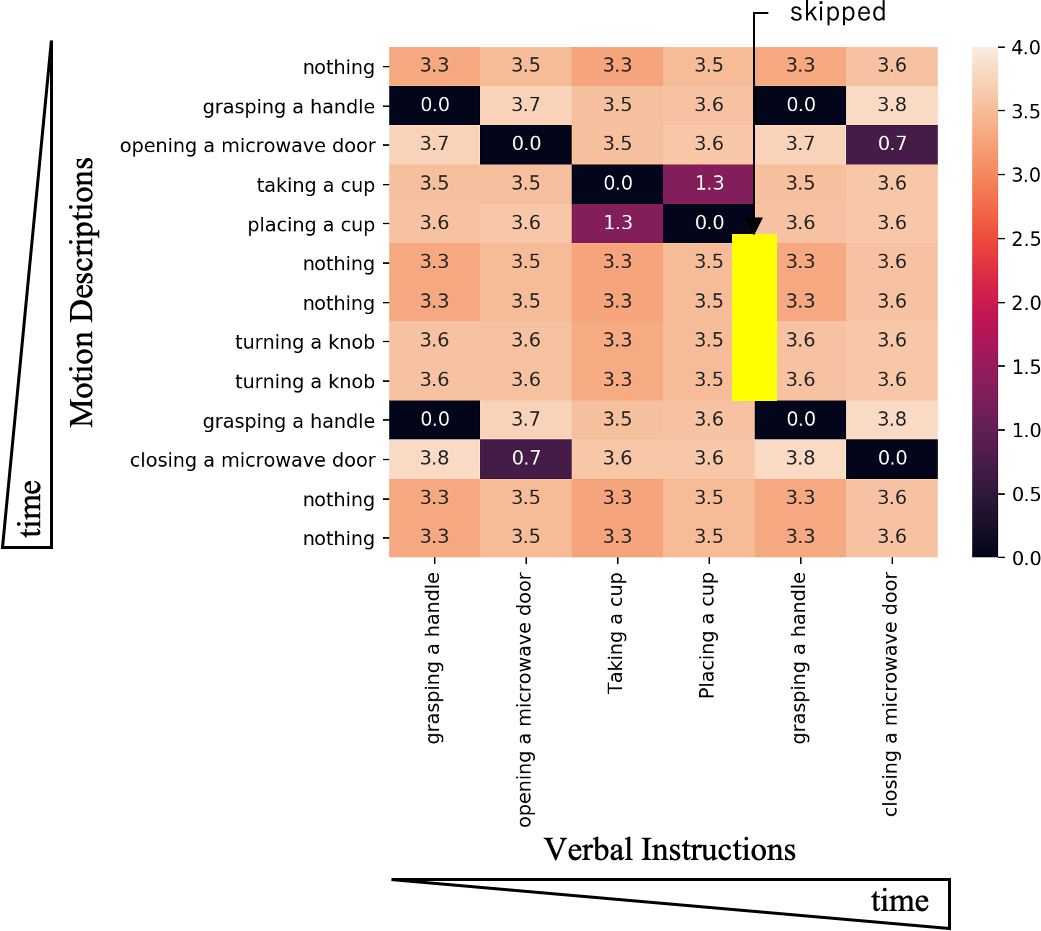}
  \caption{Word Mover's Distance for microwave operation. 
  The lower the number, the higher the match rate of sentences.
}
  \label{fig:microwave-heatmap}
\end{figure}


\subsection{Matching Evaluation}
We now evaluate the performance of the proposed matching method. Videos on our dataset have captions aligning with sections. For each video, we calculate the coincidence of matched sections by the proposed method with the ground truth sections (i.e., the sections on the testing set). Inputs are videos and sections from velocity-based video splitting. The proposed matching method outputs matched sections. 
We employed average precision (AP) at certain Intersection over Union (IoU) thresholds as the main evaluation metric. 
We calculate those IoU overlap with ground truth.
If the IoU overlap is higher than a certain threshold, we regard it as positive, otherwise negative.
Following the evaluation of TAP\cite{xiong2016cuhk},
the IoU thresholds of {0.5, 0.75, 0.95} were used.
For comparison, we prepared input sections sampled every 0.5 s.
TABLE 2 lists the AP scores.
The uniformly sampled sections' score is low 
because it is difficult to extract exact sections,
and the input sections to video captioning are not suitable for motion description.
The score of velocity splitting is high.
Note that the matching with velocity splitting method's score is high at an IoU threshold of 0.95.
This indicates that the proposed method could accurately extract the sections.

To improve the accuracy, it is necessary to improve the performance of video captioning.
The distance function also depends on Word2Vec.
The problem with Word2Vec is that vectors have no innate sense of the mechanics or functionality required to represent human actions because they are trained on text\cite{paulius2020motion}.

\subsection{Applications}
Using the proposed framework, we tested various manipulation action sequences by just watching the video.
\figref{title} illustrates an overview of the system.
We used an Azure Kinect vision sensor\cite{k4a}. 
Both RGB images and human skeleton poses were obtained through the Azure Kinect sensor with a nominal sampling rate of 30 Hz.
First, the system calculated smoothed velocity and split the input video with respect to time from the input hand positions.
Next, the system calculated optical flow images from the input video and 
motion descriptions were generated from video captioning models from RGB images and optical flow images.
Finally, matching verbal instructions with the motion descriptions, the system outputted the sections matched to each verbal instruction.


In \figref{microwave-operation}, the demonstrator is opening a microwave door and putting a cup into the microwave;
there is a time period when the demonstrator does nothing in the middle of the operation.
\figref{microwave-heatmap} shows a heatmap of WMD between the output motion descriptions and verbal instructions.
In the section where the demonstrator is not doing anything, the results of the video captioning are ``nothing" and ``turning a knob". 
Although the latter is incorrectly explained, the sections were skipped correctly owing to the high WMD with the verbal instruction.

In addition, by using the proposed framework, parameters for robot execution can be extracted.
The red-green-blue axes overlayed on\figref{microwave-operation} 
indicate that articulated object's motion parameters obtained by linear fitting and circular fitting, respectively, of the hand trajectory by extracting the section containing ``open" in the verbal instruction.
By applying our proposed method, a robot can understand action sequences and extract parameters for robots from a human demonstration video properly. 


\section{CONCLUSIONS}

Extracting exact sections and understanding action sequences from videos is important for acquiring a robot's motion knowledge.
Deep learning-based temporal action proposal and end-to-end video captioning methods lack accuracy in extracting sections.
Human motions from videos lack knowledge of whether the motions were intended.
In this paper, we proposed a new Learning-from-Observation framework, which combines video instruction and verbal instruction
to extract exact sections from videos containing action sequences.
First, we verified that hand velocity motion splitting is a good indicator for dividing daily-life actions.
Second, we created a new dataset for motion descriptions which aim to understand daily-life human actions.
Finally, we compared extracting action sequences with the velocity-based video splitting to uniformly sampled sections, and the average precision of extracting multi-step actions with the velocity-based video splitting method was clearly high at an IoU threshold of 0.95.




\bibliographystyle{ieeetr}
\bibliography{bib}

\end{document}